\documentclass{article} 
\usepackage[preprint]{colm2025_conference}

\usepackage{microtype}
\usepackage{hyperref}
\usepackage{url}
\usepackage{booktabs}
\usepackage{multirow}
\usepackage{graphicx}
\usepackage{subfloat}
\usepackage{booktabs}
\usepackage{subfig}
\usepackage{amsmath}
\usepackage{subcaption}
\usepackage[normalem]{ulem}
\usepackage{algorithm,algpseudocode}
\usepackage{amsmath}
\usepackage{amssymb}
\usepackage{xcolor}
\usepackage{lineno}
\usepackage{color}
\usepackage{colortbl}
\usepackage{amsfonts}
\usepackage{makecell}
\usepackage{paracol}
\usepackage{listings}
\usepackage{wrapfig}
\usepackage{seqsplit}
\usepackage{marvosym} 
\usepackage{textcomp}

\definecolor{myred}{RGB}{220,50,47}    
\definecolor{mygreen}{RGB}{133,153,0}  
\definecolor{darkblue}{rgb}{0, 0, 0.5}
\hypersetup{colorlinks=true, citecolor=darkblue, linkcolor=darkblue, urlcolor=darkblue}

\usepackage{enumitem}

\title{Quantitative Analysis of Performance Drop in DeepSeek Model Quantization}

\author{Enbo Zhao$^{1,2}$, Yi Shen$^{1,2}$, \Letter \  Shuming Shi$^{1,2}$, Jieyun Huang$^{1,2}$, Zhihao Chen$^{1,2}$, \\
\textbf{Ning Wang}$^{1,2}$ , \textbf{Siqi Xiao}$^{1,2}$, \textbf{Jian Zhang}$^{1,2}$,  \textbf{Kai Wang}$^{1,2}$, \Letter \ \textbf{Shiguo Lian}$^{1,2}$ \\
$^{1}$Unicom Data Intelligence, China Unicom \quad \\
$^{2}$Data Science \& Artificial Intelligence Research Institute,  China Unicom \quad \\
\Letter \ Corresponding Authors\quad \\
\texttt{\{zhaoeb7@chinaunicom.cn, sheny73@chinaunicom.cn, ssm01@hotmail.com, }\\
\texttt{ liansg@chinaunicom.cn\} }\\
}

\begin{document}

\ifcolmsubmission
\linenumbers
\fi

\maketitle










\begin{abstract}
    Recently, there is a high demand for deploying DeepSeek-R1 and V3 locally, possibly because the official service often suffers from being busy and some organizations have data privacy concerns. While single-machine deployment offers infrastructure simplicity, the models' 671B FP8 parameter configuration exceeds the practical memory limits of a standard 8-GPU machine. Quantization is a widely used technique that helps reduce model memory consumption. However, it is unclear what the performance of DeepSeek-R1 and V3 will be after being quantized. This technical report presents the first quantitative evaluation of multi-bitwidth quantization across the complete DeepSeek model spectrum. Key findings reveal that 4-bit quantization maintains little performance degradation versus FP8 while enabling single-machine deployment on standard Nvidia GPU devices. We further propose DQ3\_K\_M, a dynamic 3-bit quantization method that significantly outperforms traditional Q3\_K\_M variant on various benchmarks, which is also comparable with 4-bit quantization (Q4\_K\_M) approach in most tasks. Moreover, DQ3\_K\_M supports single-machine deployment configurations for both NVIDIA H100/A100 and Huawei 910B. Our implementation of DQ3\_K\_M is released at \url{https://github.com/UnicomAI/DeepSeek-Eval}, containing optimized 3-bit quantized variants of both DeepSeek-R1 and DeepSeek-V3.

\end{abstract}


\section{Introduction}

DeepSeek-V3 \citep{liu2024deepseek} and DeepSeek-R1 \citep{guo2025deepseek} have emerged as state-of-the-art open-source language models since their initial release, achieving top-tier performance across multiple LLM benchmarks \footnote{https://lmarena.ai/?leaderboard}. Their combination of exceptional reasoning capabilities and open accessibility has driven widespread adoption in both academic and industrial applications, resulting in a significant demand for on-premises deployment. There are two primary factors that motivate this trend:

\textbf{Service reliability}: The official DeepSeek services frequently experience capacity constraints during peak usage periods, creating operational bottlenecks for production systems.

\textbf{Data governance}: Increasing regulatory requirements and organizational privacy policies necessitate on-premises deployment for sensitive applications in scenarios such as healthcare, finance, and government sectors.

Hosting the full version of DeepSeek-V3 or DeepSeek-R1 on a single machine is appealing due to its simplicity and relatively low cost. However, 671 billion FP8 parameters exceed the device memory available on a typical single machine with 8 GPU/NPU devices (like Nvidia A100/A800/H100/H800/H20 and Huawei Ascend 910B).

Model quantization has emerged as a critical technique for efficient deployment, which helps reduce model memory consumption and enables execution on more affordable hardware configurations. For example, with 4-bit quantization (Q4), the memory cost of DeepSeek-R1's model weights (not including the KV cache and auxiliary memory for inference) is reduced from 670GB to about 370GB, which could support single-machine deployment for most popular device types. However, it is unclear what the performance of DeepSeek-R1 and V3 will be after being quantized.

In this technical report, we perform a quantitative analysis of the effectiveness of DeepSeek model quantization.
Our goal is to answer the following questions.

1. How significant is the performance drop in the quantized DeepSeek models compared to the full-precision versions?

2. Among the full model, the distilled models, and the quantized ones, which version should be deployed for a specific hardware configuration?

To investigate the answers to the above questions, we conducted a quantitative evaluation of quantized DeepSeek series of models that simultaneously examines:

\begin{itemize}
    \item Cross-Domain Consistency: Performance impacts across mathematical reasoning (MATH, AIME), code generation (MBPP, LiveCodeBench), and general knowledge (MMLU, C-Eval).
    \item Multi-Bitwidth Analysis: 2/3/4/8-bit configurations for quantization.
    \item Full-Scale Model Coverage: Comprehensive evaluation of both distilled (32B) and full-parameter (671B) DeepSeek variants.
\end{itemize}

Through rigorous evaluation, we found that the quantized DeepSeek model retains strong performance, with 4-bit quantization results often comparable to FP8 in many scenarios, demonstrating the high cost-effectiveness of quantized models. We also identified significant potential in dynamic quantization techniques.

Furthermore, by drawing insights from existing quantization techniques, we propose a dynamic 3-bit quantization method (DQ3\_K\_M) that outperforms the 3-bit quantization implementation in llama.cpp and achieves performance comparable to 4-bit quantization across multiple benchmarks.

Our contributions in this technical report are summarized as follows:

1. We conduct comprehensive evaluation of quantized DeepSeek series models. To the best of our knowledge, this is the first work in the industry to assess quantization effects on full-parameter DeepSeek models. We hope that this work can provide some reference for practitioners who aim to implement DeepSeek models in production environments.

2. We propose a dynamic 3-bit quantization method validated on full-capacity DeepSeek R1 and V3 models, which achievied strong performance. The quantized models (281G) can be conveniently deployed on a single 8 GPU/NPU device (e.g., H100 or 910B). To facilitate community use, we have open-sourced our 3-bit quantized DeepSeek models \footnote{https://github.com/UnicomAI/DeepSeek-Eval}.

In the following sections, we first review related works in Section \ref{sec:relatedwork}, followed by Section \ref{sec:method} where we introduce our proposed dynamic 3-bit quantization method (DQ3\_K\_M). Our controlled experiments, quantitative results, and practical recommendations for deployment scenarios are presented in Section \ref{sec:exp}, and finally, we conclude this work in Section \ref{sec:conclu}.
\section{Related work}
\label{sec:relatedwork}


In this section, we provide a brief overview of two techniques for LLM compression: distillation and quantization.

\subsection{Distillation} 
Knowledge distillation (KD) \citep{hinton2015distilling,romero2014fitnets}, initially proposed for developing compact yet powerful models through knowledge transfer, has evolved into a fundamental paradigm for model compression \citep{xu2024survey}. Traditional KD implementations primarily operate through  Logit-level alignment \citep{hinton2015distilling} or Intermediate feature matching \citep{romero2014fitnets}.

Recent works \citep{yang2024survey} in LLM distillation demonstrate that supervised fine-tuning (SFT) with teacher-generated outputs presents a viable alternative to conventional KD approaches. Empirical studies \citep{min2024imitate,qin2024o1,huang2024o1} have validated that this data-driven distillation paradigm enables parameter-efficient LLMs to attain competitive reasoning performance while maintaining computational tractability.

DeepSeek has assorted to the 800K training data of DeepSeek-R1 to perform SFT on the Qwen and Llama series of models, creating a series of distilled reasoning models \citep{guo2025deepseek}. We select the 32B version (DeepSeek-R1-distill-Qwen-32B) as a representative for evaluation.

\subsection{Quantization}  
Quantization constitutes a fundamental paradigm for model compression, reducing memory footprint by encoding parameters in low-precision representations \citep{gholami2022survey}. Contemporary implementations adopt two principal strategies:

\textbf{Quantization-Aware Training (QAT)}
QAT \citep{esser2019learned} integrates quantization constraints during full model retraining. While specialized adaptations like LLM-QAT\citep{liu2023llm} and EdgeQAT \citep{shen2024edgeqat} demonstrate effectiveness for moderate-scale language models, their prohibitive GPU memory demands and extended training cycles render them impractical for large-scale LLMs.

\textbf{Post-Training Quantization (PTQ)}
As a computationally efficient alternative, PTQ \citep{cai2020zeroq} converts pre-trained models to fixed-point representations without revisiting base model training. This approach requires only lightweight parameter calibration (typically <0.1\% of original training cost) through:  

\begin{equation}
    \min_{\theta} \mathbb{E}_{x \sim \mathcal{D}_{\text{calib}}} \|f_{\text{FP}}(x) - f_{\text{quant}}(\theta,x)\| ,
\end{equation}

where $\mathcal{D}_{\text{calib}}$ denotes the calibration dataset and 
$\theta$ represents the quantization scales.

PTQ techniques can be further divided into weights-only quantization and weight-activation quantization. Weights-only quantization, such as GPTQ \citep{frantar2022gptq} and SpQR\citep{dettmers2023spqr}, focuses on minimizing precision loss by adjusting weight bit-widths and applying scale transformations to preserve critical weight distributions. Weight-activation quantization \citep{xiao2023smoothquant,dettmers2022gpt3} compresses both weights and activations, utilizing techniques such as mixed-precision decomposition and channel-wise scaling to achieve an ideal compression rate with less accuracy degradation.

The quantization evaluation results in this report are all based on the weighted-only PTQ paradigm. Although preliminary efforts have been made in quantized LLM evaluation for reasoning tasks, existing studies predominantly focus on single-domain evaluations (e.g., either mathematical reasoning \citep{li2025quantization} or code generation \citep{giagnorio2025quantizing,nyamsuren2024evaluating}). Besides, current analysis \citep{liu2025quantization} about DeepSeek quantization are mainly restricted to parameter-constrained distilled variants of the DeepSeek family (less than 32B).

Unlike existing studies, our work introduced in this technical report presents the first systematic study of multi-bitwidth quantization effects across the complete DeepSeek model spectrum, including the full-parameter R1 and V3 variants (671B).

\section{Methodology}
\label{sec:method}

While existing quantization implementations for DeepSeek models demonstrate preliminary success, we posit that dynamic bit-width allocation\footnote{https://unsloth.ai/blog/deepseekr1-dynamic} based on layer importance warrants systematic exploration. Building upon the Q3 quantization baseline, we introduce adaptive precision selection guided by architectural insights. Our dynamic quantization strategy prioritizes applying higher-precision quantization to modules with fewer parameters where possible. Therefore, building upon the standard Q3 quantization provided in llama.cpp, we  implement hybrid precision by applying q6\_k or q4\_k quantization to some selected modules. Furthermore, motivated by \citep{yu2024super}'s discovery of ``super weights'' in LLMs - particularly concentrated in the  \textbf{mlp.down\_proj} layers - we observe that applying overly aggressive quantization strategies to these critical components leads to significant model performance degradation. Therefore, we implement:

\begin{itemize}
    \item q6\_k quantization for the first two \textbf{ffn\_down\_exps} layers
    \item q3\_k for subsequent layers with q4\_k inserted every fourth layer
\end{itemize}

This configuration achieves parameter distribution: 75.9\% q3\_k, 20.7\% q4\_k, and 3.4\% q6\_k within \textbf{ffn\_down\_exps} module. 

The resultant DQ3\_K\_M variant demonstrates superior memory efficiency compared to some conventional approaches (Table \ref{tab:quantization_effect}). Our implementation achieves smaller model footprint with reduced GPU memory consumption and more effective average bit-width against llama.cpp's standard Q3\_K\_M.

Quantitative performance comparisons across reasoning and generation tasks are detailed in Section \ref{sec:exp}. Complete implementation specifics of DQ3\_K\_M, including per-module quantization schemes, are provided in Appendix \ref{appendix:1}.

\begin{table*}[!ht]
\centering
\renewcommand{\arraystretch}{1.8} 
\begin{tabular}{@{}lcccccc@{}}
\toprule

\makecell{\textbf{Metric} } & 
\makecell{\textbf{Q4\_K\_M} \\ \textbf{(llama.cpp)}} & 
\makecell{\textbf{Q3\_K\_M} \\ \textbf{(llama.cpp)}} & 
\makecell{\textbf{DQ3\_K\_M} \\ \textbf{(ours)}} & 
\makecell{\textbf{Q2\_K\_L} \\ \textbf{(llama.cpp)}} & 
\makecell{\textbf{UD-Q2\_K\_XL} \\ \textbf{(Unsloth)}} & 
\\

\midrule

\textbf{Model Size} & 377G & 298G & 281G & 228G & 212G   \\
\textbf{Avg Quants} & 4.82 & 3.81 & 3.59 & 2.91 & 2.70   \\
\textbf{MU (total)} & 568GB & 487GB & 469GB & 415GB & 398GB   \\
\textbf{MU (per GPU)}& 71GB & 61GB & 59GB & 52GB & 50GB   \\

\bottomrule
\end{tabular}
\vspace{10pt} %
\caption{Comparison of resource consumption between our proposed DQ3\_K\_M and various quantization approaches provided by llama.cpp and Unsloth, using DeepSeek R1(671B) as an example. The memory usage is reported based on the maximum context length of 32K tokens. \textbf{MU} denotes Memory Usage.}
\label{tab:quantization_effect}
\end{table*}

\section{Experiments}
\label{sec:exp}

\subsection{Benchmarks} 

We conducted experiments across two categories of benchmarks \footnote{Please refer to Appendix \ref{appendix:2} for statistics of these benchmarks.}: domain-specific reasoning tasks and general capability assessments. Our reasoning benchmark suite comprises nine components :

\textbf{MATH 500} \citep{lightman2023let}: A curated subset of 500 competition-level mathematics problems from the MATH dataset \cite{hendrycks2measuring};

\textbf{AIME 2024} \footnote{https://maa.org/math-competitions/american-invitational-mathematics-examination-aime}: It features problems from the American Invitational Mathematics Examination 2024 which are specifically designed to challenge the top high school students;

\textbf{GPQA} \citep{rein2024gpqa}:  A Q\&A benchmark containing 198 multiple-choice questions spanning physics, biology, and chemistry;

\textbf{LiveCodeBench}\citep{jain2024livecodebench}: Temporal programming challenges collected from competitive coding platforms (AtCoder/LeetCode), maintaining temporal consistency (2024-08 to 2025-01) with DeepSeek-R1's evaluation protocol \citep{guo2025deepseek};

\textbf{MBPP}\citep{austin2021program}: MBPP (Mostly Basic Python Programming) is a benchmark for assessing LLM's ability to generate code for independent Python functions. It consists of 974 entry-level programming problems.

\textbf{MBPP+} \citep{liu2023your}: An enhanced variant of MBPP featuring expanded test cases and refined solution specifications.

For general capability evaluation, we adopt three established benchmarks:

\textbf{MMLU} \citep{hendrycks2020measuring} (Massive Multitask Language Understanding): widely used benchmark for LLM evaluation contains diverse questions across 57 academic subjects.

\textbf{CMMLU} \citep{li2023cmmlu} (Chinese Massive Multitask Language Understanding):
11582 Chinese questions spanning STEM and humanities.

\textbf{C-Eval} \citep{huang2023c}: 12342 challenging Chinese exam-style questions.

\begin{table*}
\centering
\renewcommand{\arraystretch}{1.8} 
\begin{tabular}{@{}lccccccc@{}}
\toprule

\multirow{2}{*}[0.5em]{ \small \textbf{DeepSeek-R1}}  & 
\makecell{\small \textbf{FP8} \\ \small \textbf{(Reported)}} & 
\makecell{\small \textbf{FP8} \\ \small \textbf{(Official API)}} & 
\makecell{\small \textbf{Q4\_K\_M} \\ \small \textbf{(llama.cpp)}} & 
\makecell{\small \textbf{Q3\_K\_M} \\ \small \textbf{(llama.cpp)}} & 
\makecell{\small \textbf{UD-Q2\_K\_XL} \\ \small \textbf{(Unsloth)}} &
\makecell{\small \textbf{DQ3\_K\_M} \\ \small \textbf{(Ours)} }\\

\midrule
\small \textbf{AIME 2024} & 79.8 & \makecell[c]{77.53 \\ \small ($\pm$2.97)}  & \makecell[c]{75.43 \\ \small ($\pm$3.07)}  &  \makecell[c]{ 72.50 \\ \small ($\pm$6.11)}   &  \makecell[c]{ 75.83 \\ \small ($\pm$5.83)} &  \makecell[c]{ 75.41 \\ \small ($\pm$4.69)} \\

\small \textbf{MATH 500} & 97.3 & \makecell[c]{95.45 \\ \small ($\pm$0.82)}  & \makecell[c]{95.55 \\ \small ($\pm$0.44)}  &  \makecell[c]{ 94.15 \\ \small ($\pm$0.68)}   &  \makecell[c]{ 95.25 \\ \small ($\pm$0.44)} &  \makecell[c]{ 95.35 \\ \small ($\pm$0.50)} \\

\small \textbf{GPQA} & 71.5 & \makecell[c]{69.58 \\ \small ($\pm$1.65)}  & \makecell[c]{69.95 \\ \small ($\pm$1.85)}  &  \makecell[c]{ 65.80 \\ \small ($\pm$2.30)}   &  \makecell[c]{ 68.93 \\ \small ($\pm$1.55)} &  \makecell[c]{ 68.95 \\ \small ($\pm$0.65)} \\

\small \textbf{MBPP} & - & \makecell[c]{92.60 \\ \small ($\pm$0.80)}  & \makecell[c]{91.60 \\ \small ($\pm$2.00)}  &  \makecell[c]{ 90.43 \\ \small ($\pm$0.88)}   &  \makecell[c]{ 92.93 \\ \small ($\pm$0.24)} &  \makecell[c]{ 92.80 \\ \small ($\pm$0.70)} \\

\small \textbf{MBPP+} & - & \makecell[c]{78.35 \\ \small ($\pm$1.06)}  & \makecell[c]{76.70 \\ \small ($\pm$1.85)}  &  \makecell[c]{ 76.75\\ \small ($\pm$0.88)}   &  \makecell[c]{ 78.33 \\ \small ($\pm$0.91)} &  \makecell[c]{ 78.60 \\ \small ($\pm$1.01)} \\

\small \textbf{LiveCodeBench} & 65.9 & \makecell[c]{64.16 \\ \small ($\pm$1.51)}  & \makecell[c]{62.41 \\ \small ($\pm$2.27)}  &  \makecell[c]{ 61.95\\ \small ($\pm$1.66)}   &  \makecell[c]{ 61.40 \\ \small ($\pm$1.59)} &  \makecell[c]{ 63.15 \\ \small ($\pm$1.06)} \\

\small \textbf{MMLU} & 90.8 & \textbf{90.99} & 90.14 & 89.87 & 89.72  & 91.03 \\
\small \textbf{CMMLU} & - & 90.37 & \textbf{90.42} & 89.85 & 89.61 & 90.17 \\
\small \textbf{C-Eval} & 91.8 & 92.20 & 92.10 & 91.60 & 91.70 & 91.80\\

\midrule
\small \textbf{Average} & - & 83.48 & 82.70 & 81.44 & 82.63 & 83.03  \\
\small \textbf{Weighted avg.} & - & 85.82 & 85.24 & 84.28 & 85.02 & 85.53  \\
\small \textbf{Accuracy drop} & - & - & 0.68\% & 1.80\% & 0.94\% & \textbf{0.34}\%  \\

\bottomrule
\end{tabular}
\vspace{2pt} %
\caption{Main results of DeepSeek-R1 on various benchmarks. \textbf{Accuracy drop} refers to the relative percentage decrease in average score against the results from FP8 (Official API).}
\label{tab:performance_R1}
\end{table*}

\begin{table*}
\centering
\renewcommand{\arraystretch}{1.8} 
\begin{tabular}{@{}lccccccc@{}}
\toprule

\multirow{2}{*}[0.5em]{\small \textbf{DeepSeek-V3}}  & 
\makecell{\small \textbf{FP8} \\ \small \textbf{(Reported)}} & 
\makecell{\small \textbf{FP8} \\ \small \textbf{(Tencent API)}} & 
\makecell{\small \textbf{Q4\_K\_M} \\ \small \textbf{(llama.cpp)}} & 
\makecell{\small \textbf{Q3\_K\_M} \\ \small \textbf{(llama.cpp)}} & 
\makecell{\small \textbf{Q2\_K\_L} \\ \small \textbf{(llama.cpp)}} &
\makecell{\small \textbf{DQ3\_K\_M} \\ \small \textbf{(Ours)}} \\

\midrule
\small \textbf{AIME 2024} & 39.2 & \makecell[c]{38.34 \\ \small ($\pm$2.52)}  & \makecell[c]{41.66 \\ \small ($\pm$4.72)}  &  \makecell[c]{ 38.73 \\ \small ($\pm$4.70)}   &  \makecell[c]{ 15.41 \\ \small ($\pm$3.55)} &  \makecell[c]{ 39.16 \\ \small ($\pm$4.97)} \\

\small \textbf{MATH 500} & 90.2 & \makecell[c]{ 89.85 \\ \small ($\pm$0.30)} & \makecell[c]{90.55 \\ \small ($\pm$0.44)}  & \makecell[c]{89.05 \\ \small ($\pm$1.27)}  &  \makecell[c]{ 77.30 \\ \small ($\pm$0.66)}   &  \makecell[c]{ 89.65 \\ \small ($\pm$0.98)} \\


\small \textbf{GPQA} & 59.1 & \makecell[c]{52.23 \\ \small ($\pm$3.44)}  & \makecell[c]{51.95 \\ \small ($\pm$2.64)}  &  \makecell[c]{ 52.13 \\ \small ($\pm$1.25)}   &  \makecell[c]{ 43.65 \\ \small ($\pm$1.32)} &  \makecell[c]{ 52.38\\ \small ($\pm$1.31)} \\


\small \textbf{MBPP} & - & \makecell[c]{87.75 \\ \small ($\pm$0.61)}  & \makecell[c]{87.18 \\ \small ($\pm$0.70)}  &  \makecell[c]{ 88.55 \\ \small ($\pm$0.90)}   &  \makecell[c]{ 81.10 \\ \small ($\pm$1.55)} &  \makecell[c]{ 89.38 \\ \small ($\pm$0.35)} \\


\small \textbf{MBPP+} & - & \makecell[c]{73.35 \\ \small ($\pm$1.21)}  & \makecell[c]{72.90 \\ \small ($\pm$0.66)}  &  \makecell[c]{ 73.08\\ \small ($\pm$1.31)}   &  \makecell[c]{ 67.83 \\ \small ($\pm$1.09)} &  \makecell[c]{ 74.78 \\ \small ($\pm$0.56)} \\


\small \textbf{LiveCodeBench} & 36.2 & \makecell[c]{36.21 \\ \small ($\pm$0.47)}  & \makecell[c]{37.40 \\ \small ($\pm$1.32)}  &  \makecell[c]{ 36.21\\ \small ($\pm$2.03)}   &  \makecell[c]{ 29.14 \\ \small ($\pm$0.92)} &  \makecell[c]{ 36.76 \\ \small ($\pm$0.67)} \\


\small \textbf{MMLU} & 88.5 & 88.06 & 88.09 & 87.31 & 84.25  & 87.87 \\
\small \textbf{CMMLU} & - & 81.57 & 82.68 & 80.69 & 77.32 & 81.07 \\
\small \textbf{C-Eval} & 86.5& 83.10 & 82.90 & 82.60 & 77.60 & 83.40  \\

\midrule
\small \textbf{Average} & - & 70.05 & 70.59 & 69.82 & 61.51 & 70.47  \\
\small \textbf{Weighted avg.} & - & 75.45 & 75.79 & 75.06 & 68.73 & 75.73  \\
\small \textbf{Accuracy drop} & - & - & \textbf{0} & 0.52\% & 8.91\% & \textbf{0} \\

\bottomrule
\end{tabular}
\vspace{1pt} %
\caption{Quantization results of DeepSeek-V3 on various benchmarks. }
\label{tab:performance_V3}
\end{table*}


\begin{table*}
\centering
\renewcommand{\arraystretch}{1.8} 
\begin{tabular}{@{}lccccccc@{}}
\toprule

\makecell{\small \textbf{DeepSeek-V3} \\ \small  \textbf{0324}} &
\makecell{\small \textbf{FP8} \\ \small \textbf{(Official API)}} & 
\makecell{\small \textbf{Q4\_K\_M} \\ \small \textbf{(llama.cpp)}} & 
\makecell{\small \textbf{Q3\_K\_M} \\ \small \textbf{(llama.cpp)}} & 
\makecell{\small \textbf{Q2\_K\_L} \\ \small \textbf{(llama.cpp)}} &
\makecell{\small \textbf{DQ3\_K\_M} \\ \small \textbf{(Ours)}} &
\makecell{\small \textbf{Q4\_K} \\ } &
\makecell{\small \textbf{Q3\_K} \\ } \\

\midrule
\small \textbf{AIME 2024} & \makecell[c]{57.9 \\ \small ($\pm$4.34)}   & \makecell[c]{53.3 \\ \small ($\pm$3.10)}  &  \makecell[c]{ 54.57 \\ \small ($\pm$6.14)}   &\makecell[c]{31.25 \\ \small ($\pm$3.04)} & \makecell[c]{ 57.09 \\ \small ($\pm$5.16)} &  \makecell[c]{ 59.18 \\ \small ($\pm$7.91)} & \makecell[c]{52.51 \\ \small ($\pm$5.29)} \\

\small \textbf{MATH 500} & \makecell[c]{93.25 \\ \small ($\pm$0.91)}   & \makecell[c]{93.25 \\ \small ($\pm$0.47)}  &  \makecell[c]{ 92.50 \\ \small ($\pm$0.96)}   &\makecell[c]{85.30 \\ \small ($\pm$0.68)} & \makecell[c]{ 93.55 \\ \small ($\pm$0.25)} &  \makecell[c]{ 93.0 \\ \small ($\pm$1.06)} & \makecell[c]{91.65 \\ \small ($\pm$1.34)} \\


\small \textbf{GPQA} & \makecell[c]{60.48 \\ \small ($\pm$1.38)}   & \makecell[c]{59.10 \\ \small ($\pm$1.73)}  &  \makecell[c]{ 59.98 \\ \small ($\pm$0.95)}   &\makecell[c]{46.75 \\ \small ($\pm$0.96)} & \makecell[c]{ 60.23 \\ \small ($\pm$1.11)} &  \makecell[c]{ 56.20 \\ \small ($\pm$2.15)} & \makecell[c]{61.35 \\ \small ($\pm$2.60)} \\


\small \textbf{MBPP} & \makecell[c]{89.03 \\ \small ($\pm$0.53)}   & \makecell[c]{88.63 \\ \small ($\pm$0.56)}  &  \makecell[c]{ 88.10 \\ \small ($\pm$0.41)}   &\makecell[c]{82.93 \\ \small ($\pm$1.04)} & \makecell[c]{ 89.50 \\ \small ($\pm$0.24)} &  \makecell[c]{ 88.43 \\ \small ($\pm$1.87)} & \makecell[c]{87.78 \\ \small ($\pm$1.11)} \\


\small \textbf{MBPP+} & \makecell[c]{74.73 \\ \small ($\pm$0.48)}   & \makecell[c]{74.40 \\ \small ($\pm$0.74)}  &  \makecell[c]{ 73.08 \\ \small ($\pm$0.30)}   &\makecell[c]{68.98 \\ \small ($\pm$1.00)} & \makecell[c]{ 75.63 \\ \small ($\pm$0.54)} &  \makecell[c]{ 73.33 \\ \small ($\pm$2.13)} & \makecell[c]{73.30 \\ \small ($\pm$1.06)} \\


\small \textbf{LiveCodeBench} & \makecell[c]{49.73 \\ \small ($\pm$1.26)}   & \makecell[c]{47.88 \\ \small ($\pm$1.21)}  &  \makecell[c]{ 46.23 \\ \small ($\pm$0.46)}   &\makecell[c]{36.95 \\ \small ($\pm$0.70)} & \makecell[c]{ 47.89 \\ \small ($\pm$0.35)} &  \makecell[c]{ 47.79 \\ \small ($\pm$1.04)} & \makecell[c]{44.95 \\ \small ($\pm$0.97)} \\


\small \textbf{MMLU} & 89.08 &  88.71 & 88.47 & 85.59  & 88.93 &88.73 & 88.57 \\
\small \textbf{CMMLU} & 86.13 &  86.13 & 85.28 & 81.57 & 85.99 & 85.96 &  84.84\\
\small \textbf{C-Eval} & 89.60 & 89.10 & 88 .90 & 73.60 & 89.10 & 89.00 &  88.50\\

\midrule
\small \textbf{Average} &  76.66 & 75.62 & 75.24 & 65.88  & 76.43 & 75.74 & 74.83\\
\small \textbf{Weighted avg.} & 80.70 &  80.04 & 79.56 & 71.49 & 80.50  & 79.81 & 79.29  \\
\small \textbf{Accuracy drop} & - & 1.35\% & 1.85\% & 14.66\% & 0.30\% &1.20\% & 2.39\% \\

\bottomrule
\end{tabular}
\vspace{1pt} %
\caption{Quantization results of DeepSeek-V3-0324 on various benchmarks. }
\label{tab:performance_V3_0324}
\end{table*}


\begin{table*}
\centering
\renewcommand{\arraystretch}{1.8} 
\begin{tabular}{@{}lcccccc@{}}
\toprule

\makecell{\small \textbf{DeepSeek-R1} \\ \small  \textbf{distill-Qwen-32B}} & 
\makecell{\small \textbf{BF16} \\ \small \textbf{(Reported)}} & 
\makecell{\small \textbf{BF16} \\ \small \textbf{(Local Evaluation)}} & 
\makecell{\small \textbf{Q8\_0} \\ \small \textbf{(llama.cpp)}} & 
\makecell{\small \textbf{Q4\_K\_M} \\ \small \textbf{(llama.cpp)}} & 
\makecell{\small \textbf{Q3\_K\_M} \\ \small \textbf{(llama.cpp)}} & 
\\

\midrule
\small \textbf{AIME 2024} & 72.6 & \makecell[c]{69.59 \\ \small ($\pm$2.75)} & \makecell[c]{71.68 \\ \small ($\pm$4.71)} & \makecell[c]{70.40 \\ \small ($\pm$7.66)}  & \makecell[c]{71.24 \\ \small ($\pm$6.66)}  \\


\small \textbf{MATH 500} & 94.3 & \makecell[c]{93.65 \\ \small ($\pm$0.41)} & \makecell[c]{93.10 \\ \small ($\pm$0.42)} & \makecell[c]{93.90 \\ \small ($\pm$0.53)}  & \makecell[c]{93.50 \\ \small (0.38)}  \\


\small \textbf{GPQA} & 62.1 & \makecell[c]{61.85 \\ \small ($\pm$2.18)} & \makecell[c]{58.85 \\ \small ($\pm$2.75)} & \makecell[c]{62.00 \\ \small ($\pm$4.54)}  & \makecell[c]{60.20 \\ \small ($\pm$1.95)}  \\


\small \textbf{LiveCodeBench} & 57.2 & \makecell[c]{57.08 \\ \small ($\pm$1.01)} & \makecell[c]{57.59 \\ \small ($\pm$1.17)} & \makecell[c]{56.85 \\ \small ($\pm$2.87)}  & \makecell[c]{55.20 \\ \small ($\pm$1.74)}  \\


\small \textbf{MBPP} & - & \makecell[c]{89.35 \\ \small ($\pm$0.42)} & \makecell[c]{89.35 \\ \small ($\pm$0.73)} & \makecell[c]{89.73 \\ \small ($\pm$1.20)}  & \makecell[c]{88.93 \\ \small ($\pm$0.64)}  \\


\small \textbf{MBPP+} & - & \makecell[c]{75.43 \\ \small ($\pm$0.91)} & \makecell[c]{75.45 \\ \small ($\pm$1.18)} & \makecell[c]{75.53 \\ \small ($\pm$1.04)}  & \makecell[c]{75.38 \\ \small ($\pm$1.30)}  \\


\small \textbf{MMLU} & - & 82.15 & 82.15 & 82.37 & 82.17   \\
\small \textbf{CMMLU} & - & 83.91 & 83.97 & 83.57 & 83.34  \\
\small \textbf{C-Eval} & - & 87.0 & 86.7 & 86.8 &  86.2  \\

\midrule
\small \textbf{Average} & - & 77.78 & 77.65 & 77.91 & 77.35   \\
\small \textbf{Weighted avg.} & - & 79.94 & 79.71 & 79.97 & 79.40   \\
\small \textbf{Accuracy drop} & - & - & 0.29\% & 0 & 0.68\%   \\

\bottomrule
\end{tabular}
\vspace{2pt} %
\caption{Results of DeepSeek-R1-distill-Qwen-32B on various benchmarks. \textbf{Accuracy drop} denotes the relative  decrease in average score against the results from BF16.}
\label{tab:performance_32B}
\end{table*}

\subsection{Experimental Setting} 
We evaluate the performance of quantized models from three original models (DeepSeek-V3, DeepSeek-R1 and DeepSeek-R1-distill-Qwen-32B) across multiple bit-width configurations on the aforementioned benchmarks. Our post-training quantization (PTQ) implementation leverages two established frameworks:

\begin{enumerate}
    \item llama.cpp\footnote{https://github.com/ggml-org/llama.cpp} for 4-bit (Q4\_K\_M), 3-bit (Q3\_K\_M), 2-bit (Q2\_K), and 8-bit (Q8\_0) configurations
    \item Unsloth\footnote{https://unsloth.ai/blog/deepseekr1-dynamic} for specialized dynamic 2-bit quantization (Q2\_K\_XL)
\end{enumerate}

\textbf{Quantization Setting}

The quantization configurations shared by all models include:

    \begin{itemize}
        \item 4-bit: Q4\_K\_M (llama.cpp)
        \item 3-bit: Q3\_K\_M (llama.cpp)
    \end{itemize}

Model-specific quantization implementations:

    \begin{itemize}
        \item DeepSeek-V3 2-bit: Standard Q2\_K (llama.cpp)
        \item DeepSeek-R1 2-bit: Large-scale UD-Q2\_K\_XL  (unsloth)
        \item DeepSeek-distill-Qwen-32B 8-bit: Q8\_0  (llama.cpp)
    \end{itemize}

For DeepSeek-R1 and DeepSeek-V3, we also conduct additional performance evaluations of our proposed Q3 quantization implementation (DQ3\_K\_M).

\textbf{Decoding Configuration}

All quantized models were configured with a maximum generation length fixed at 32,768 tokens. We used a temperature of 0.6 and a top-p value of 0.95. We implemented differentiated decoding strategies across benchmark categories:

\begin{enumerate}
    \item For small benchmarks (MATH 500, GPQA, LiveCodeBench, etc.), we employ rigorous statistical sampling: generating 4 independent responses per query and compute mean scores across samples to mitigate variance. Since AIME 2024 only contains 30 questions, we sampled 8 responses for each question.
    
    \item For large benchmarks (MMLU, CMMLU, and C-Eval), we adopt a single inference pass per question, as we observe relatively stable results on these benchmarks.
\end{enumerate}

\subsection{Main Results} 
The evaluation results for DeepSeek-R1, DeepSeek-V3 and DeepSeek-R1-distill-Qwen-32B are shown in Table \ref{tab:performance_R1}, \ref{tab:performance_V3}, and \ref{tab:performance_32B}, respectively. We present the official evaluation results reported in \citep{guo2025deepseek}, official deepSeek API\footnote{https://api-docs.deepseek.com/}  invocation outcomes, and performance metrics of different quantized model variants. For multi-sampling results, we report mean values with corresponding standard deviations (in parentheses). Notably, due to the official DeepSeek V3 API update on March 24, 2025, we substituted it with Tencent's DeepSeek V3 API \footnote{https://cloud.tencent.com/document/product/1772/115963} in Table 2 to ensure comparability.

\textbf{DeepSeek-R1} \\
Table \ref{tab:performance_R1} demonstrates the impact of various quantization methods on DeepSeek-R1's performance across multiple benchmarks. While the official FP8 demonstrates superior overall performance, Q4\_K\_M quantization methods exhibit surprisingly competitive results. Across most reasoning benchmarks, Q4\_K\_M shows no significant performance degradation, with its metrics on MATH 500 and GPQA even  marginally outperforming the official FP8 API implementation. The performance on general capability benchmarks remains stable across all variants, suggesting core semantic representations withstand quantization.

The proposed DQ3\_K\_M approach attaining an average score of 83.03 that surpasses standard 3-bit implementations and closely matches the performance of Q4\_K\_M. These results, in conjunction with Table \ref{tab:quantization_effect}, demonstrate that DQ3\_K\_M achieves superior efficiency without compromising capability. The performance stability of DQ3\_K\_M also proves particularly noteworthy, achieving lowest standard deviation in scientific QA (0.65 on GPQA vs Q3\_K\_M's 2.30) and coding benchmarks (1.06 on LiveCodeBench vs Q4\_K\_M's 2.27). We also noticed that dynamic quantized 2-bit model (UD-Q2\_K\_XL) outperforms standard 3-bit quantization (Q3\_K\_M) across multiple benchmarks, further validates the fundamental benefits of dynamic quantization.

\textbf{DeepSeek-V3}   

As evidenced in Table \ref{tab:performance_V3}, the DeepSeek-V3 model exhibits similar quantization characteristics to DeepSeek-R1 in general. Through comparison of standard quantization variants from llama.cpp, we could find that Q3\_K\_M performs slightly worse than Q4\_K\_M (with weighted average of 75.06 vs. 75.79). Approach Q2\_K\_M exhibits severe performance degradation in all evaluated benchmarks (61.51 vs. 70.05 on average when compared with FP8). This empirically validates the inevitable accuracy-compression trade-off in LLM quantization, where aggressive bit-width reduction fundamentally disrupts model capabilities.
With a weighted average benchmark score of 75.73, our proposed dynamic 3-bit quantization (DQ3\_K\_M) performs similarly with Q4\_K\_M (75.79) and FP8 (75.45). These results again demonstrate the effectiveness of our newly proposed quantization approach.






\textbf{DeepSeek-V3-0324}   

The evaluation results in Table \ref{tab:performance_V3_0324} reveal that DeepSeek-V3-0324 maintains strong performance when using our proposed DQ3\_K\_M method. It achieves near-lossless compression (average drop: 0.30\%), outperforming the FP8 baseline on MATH 500 (93.55 vs. 93.25). Extreme 2-bit quantization (Q2\_K\_L) causes severe degradation (-14.66\% average drop), particularly in knowledge-intensive tasks (e.g., C-Eval: 73.60 vs. 89.60). Our method consistently surpasses llama.cpp’s 3-bit variant (Q3\_K\_M) at the same bit-width and even its 4-bit implementation (Q4\_K\_M). We also developed fully quantized versions at 3-bit (Q3\_K) and 4-bit (Q4\_K) precision. DQ3\_K\_M outperforms both alternatives in average performance metrics. These results demonstrate that DQ3\_K\_M enables efficient deployment with minimal performance trade-offs.

\textbf{DeepSeek-R1-distill-Qwen-32B} 

 As reported in Table \ref{tab:performance_32B}, our systematic evaluation of DeepSeek-R1-distill-Qwen-32B reveals that 4-bit quantization (Q4\_K\_M) achieves optimal performance preservation, maintaining the performance of the original BF16 format across diverse benchmarks while reducing memory requirements significantly. This configuration demonstrates particular robustness in mathematical reasoning (MATH: 93.90 vs 93.65 local BF16) and scientific QA (GPQA: 62.00 vs 61.85), despite exhibiting higher standard deviation in complex tasks ($\sigma=7.66$ for AIME 2024).
For code generation tasks, MBPP and MBPP+ show remarkable quantization resilience performance variation across bit-widths. However, LiveCodeBench shows sensitivity to aggressive quantization (Q3\_K\_M: 55.20 vs Q8\_0: 57.08). This may be due to the relatively high difficulty of LiveCodeBench. In addition, consistent performance preservation ($\varDelta < 0.8\% $) across MMLU/CMMLU/C-Eval shows exceptional robustness of different bit-widths, which demonstrates that quantization preserves the general language understanding capabilities of the distillation model.

\begin{table*}[!ht]
\centering
\renewcommand{\arraystretch}{1.8} 
\begin{tabular}{@{}lcccccc@{}}
\toprule

\makecell{\textbf{Metric} } & 
\makecell{\textbf{Q4\_K\_M} \\ \textbf{(llama.cpp)}} & 
\makecell{\textbf{Q3\_K\_M} \\ \textbf{(llama.cpp)}} & 
\makecell{\textbf{DQ3\_K\_M} \\ \textbf{(Ours)}} & 
\makecell{\textbf{Q2\_K\_L} \\ \textbf{(llama.cpp)}} & 
\makecell{\textbf{UD-Q2\_K\_XL} \\ \textbf{(Unsloth)}} & 
\\

\midrule

\textbf{Avg. Score (V3)} & 75.79 & 75.06 & 75.73 & 68.73 & -   \\
\textbf{Avg. Score (R1)} & 85.24 & 84.28 & 85.53 & - & 85.02   \\
\textbf{MU (total)} & 568GB & 487GB & 469GB & 415GB & 398GB   \\
\textbf{MU (per GPU)} & 71GB & 61GB & 59GB & 52GB & 50GB   \\

\bottomrule
\end{tabular}
\vspace{10pt} %
\caption{
Comparison among various quantization approaches in terms of accuracy and memory usage. Memory usage (\textbf{MU}) is reported based on the maximum context length of 32K tokens.}
\label{tab:trade-off}
\end{table*}

\subsection{Recommendations for Different Devices}

Based on the statistical analysis in Table \ref{tab:trade-off}, we conclude that for full-parameter R1 and V3 models, 4-bit quantization (Q4\_K\_M) and our DQ3\_K\_M achieve optimal cost-performance ratio under NVIDIA-based single-machine deployments (e.g., 80GB VRAM per A100/A800/H100/H800/H20 GPU). However, Q4 typically exceeds the VRAM constraints of Huawei Ascend 910B single-node configurations (64GB per NPU), whereas DQ3\_K\_M satisfies both NVIDIA H100 and Ascend 910B configuration. Compared to other quantization variants, our dynamic 3-bit quantization DQ3\_K\_M achieves a favorable performance-resource trade-off.

\section{Conclusion}
\label{sec:conclu}

This work presents the first systematic evaluation of multi-bitwidth quantization for various deepseek models including 671B-scale, employing a comprehensive analysis across multi-domain benchmarks. Our findings demonstrate that standard 4-bit quantization (Q4) exhibits minimal performance degradation versus FP8 while significantly reducing memory requirements. We further introduce DQ3\_K\_M, a dynamic Q3 quantization method with higher memory compression ratio that surpasses the current state-of-the-art Q3\_K\_M implementation in llama.cpp, achieves 1.48\% and  0.89\% improvement on average on R1 and V3, respectively. This study establishes that careful quantization design can retain the vast majority of the original model's capabilities with only tiny performance loss while enabling cost-effective deployment on single a single machine with 8 GPU devices. We will explore more efficient quantization techniques in the future.

\bibliography{colm2024_conference}

\begin{thebibliography}{33}
\providecommand{\natexlab}[1]{#1}
\providecommand{\url}[1]{\texttt{#1}}
\expandafter\ifx\csname urlstyle\endcsname\relax
  \providecommand{\doi}[1]{doi: #1}\else
  \providecommand{\doi}{doi: \begingroup \urlstyle{rm}\Url}\fi

\bibitem[Austin et~al.(2021)Austin, Odena, Nye, Bosma, Michalewski, Dohan, Jiang, Cai, Terry, Le, et~al.]{austin2021program}
Jacob Austin, Augustus Odena, Maxwell Nye, Maarten Bosma, Henryk Michalewski, David Dohan, Ellen Jiang, Carrie Cai, Michael Terry, Quoc Le, et~al.
\newblock Program synthesis with large language models.
\newblock \emph{arXiv e-prints}, pp.\  arXiv--2108, 2021.

\bibitem[Cai et~al.(2020)Cai, Yao, Dong, Gholami, Mahoney, and Keutzer]{cai2020zeroq}
Yaohui Cai, Zhewei Yao, Zhen Dong, Amir Gholami, Michael~W Mahoney, and Kurt Keutzer.
\newblock Zeroq: A novel zero shot quantization framework.
\newblock In \emph{Proceedings of the IEEE/CVF conference on computer vision and pattern recognition}, pp.\  13169--13178, 2020.

\bibitem[Dettmers et~al.(2022)Dettmers, Lewis, Belkada, and Zettlemoyer]{dettmers2022gpt3}
Tim Dettmers, Mike Lewis, Younes Belkada, and Luke Zettlemoyer.
\newblock Gpt3. int8 (): 8-bit matrix multiplication for transformers at scale.
\newblock \emph{Advances in neural information processing systems}, 35:\penalty0 30318--30332, 2022.

\bibitem[Dettmers et~al.(2023)Dettmers, Svirschevski, Egiazarian, Kuznedelev, Frantar, Ashkboos, Borzunov, Hoefler, and Alistarh]{dettmers2023spqr}
Tim Dettmers, Ruslan Svirschevski, Vage Egiazarian, Denis Kuznedelev, Elias Frantar, Saleh Ashkboos, Alexander Borzunov, Torsten Hoefler, and Dan Alistarh.
\newblock Spqr: A sparse-quantized representation for near-lossless llm weight compression.
\newblock \emph{arXiv preprint arXiv:2306.03078}, 2023.

\bibitem[Esser et~al.(2019)Esser, McKinstry, Bablani, Appuswamy, and Modha]{esser2019learned}
Steven~K Esser, Jeffrey~L McKinstry, Deepika Bablani, Rathinakumar Appuswamy, and Dharmendra~S Modha.
\newblock Learned step size quantization.
\newblock \emph{arXiv preprint arXiv:1902.08153}, 2019.

\bibitem[Frantar et~al.(2022)Frantar, Ashkboos, Hoefler, and Alistarh]{frantar2022gptq}
Elias Frantar, Saleh Ashkboos, Torsten Hoefler, and Dan Alistarh.
\newblock Gptq: Accurate post-training quantization for generative pre-trained transformers.
\newblock \emph{arXiv preprint arXiv:2210.17323}, 2022.

\bibitem[Gholami et~al.(2022)Gholami, Kim, Dong, Yao, Mahoney, and Keutzer]{gholami2022survey}
Amir Gholami, Sehoon Kim, Zhen Dong, Zhewei Yao, Michael~W Mahoney, and Kurt Keutzer.
\newblock A survey of quantization methods for efficient neural network inference.
\newblock In \emph{Low-power computer vision}, pp.\  291--326. Chapman and Hall/CRC, 2022.

\bibitem[Giagnorio et~al.(2025)Giagnorio, Mastropaolo, Afrin, Di~Penta, and Bavota]{giagnorio2025quantizing}
Alessandro Giagnorio, Antonio Mastropaolo, Saima Afrin, Massimiliano Di~Penta, and Gabriele Bavota.
\newblock Quantizing large language models for code generation: A differentiated replication.
\newblock \emph{arXiv preprint arXiv:2503.07103}, 2025.

\bibitem[Guo et~al.(2025)Guo, Yang, Zhang, Song, Zhang, Xu, Zhu, Ma, Wang, Bi, et~al.]{guo2025deepseek}
Daya Guo, Dejian Yang, Haowei Zhang, Junxiao Song, Ruoyu Zhang, Runxin Xu, Qihao Zhu, Shirong Ma, Peiyi Wang, Xiao Bi, et~al.
\newblock Deepseek-r1: Incentivizing reasoning capability in llms via reinforcement learning.
\newblock \emph{arXiv preprint arXiv:2501.12948}, 2025.

\bibitem[Hendrycks et~al.(2020)Hendrycks, Burns, Basart, Zou, Mazeika, Song, and Steinhardt]{hendrycks2020measuring}
Dan Hendrycks, Collin Burns, Steven Basart, Andy Zou, Mantas Mazeika, Dawn Song, and Jacob Steinhardt.
\newblock Measuring massive multitask language understanding.
\newblock \emph{arXiv preprint arXiv:2009.03300}, 2020.

\bibitem[Hendrycks et~al.(2021)Hendrycks, Burns, Kadavath, Arora, Basart, Tang, Song, and Steinhardt]{hendrycks2measuring}
Dan Hendrycks, Collin Burns, Saurav Kadavath, Akul Arora, Steven Basart, Eric Tang, Dawn Song, and Jacob Steinhardt.
\newblock Measuring mathematical problem solving with the math dataset.
\newblock In \emph{Thirty-fifth Conference on Neural Information Processing Systems Datasets and Benchmarks Track (Round 2)}, 2021.

\bibitem[Hinton et~al.(2015)Hinton, Vinyals, and Dean]{hinton2015distilling}
Geoffrey Hinton, Oriol Vinyals, and Jeff Dean.
\newblock Distilling the knowledge in a neural network.
\newblock \emph{arXiv preprint arXiv:1503.02531}, 2015.

\bibitem[Huang et~al.(2023)Huang, Bai, Zhu, Zhang, Zhang, Su, Liu, Lv, Zhang, Fu, et~al.]{huang2023c}
Yuzhen Huang, Yuzhuo Bai, Zhihao Zhu, Junlei Zhang, Jinghan Zhang, Tangjun Su, Junteng Liu, Chuancheng Lv, Yikai Zhang, Yao Fu, et~al.
\newblock C-eval: A multi-level multi-discipline chinese evaluation suite for foundation models.
\newblock \emph{Advances in Neural Information Processing Systems}, 36:\penalty0 62991--63010, 2023.

\bibitem[Huang et~al.(2024)Huang, Zou, Li, Liu, Zheng, Chern, Xia, Qin, Yuan, and Liu]{huang2024o1}
Zhen Huang, Haoyang Zou, Xuefeng Li, Yixiu Liu, Yuxiang Zheng, Ethan Chern, Shijie Xia, Yiwei Qin, Weizhe Yuan, and Pengfei Liu.
\newblock O1 replication journey--part 2: Surpassing o1-preview through simple distillation, big progress or bitter lesson?
\newblock \emph{arXiv preprint arXiv:2411.16489}, 2024.

\bibitem[Jain et~al.(2024)Jain, Han, Gu, Li, Yan, Zhang, Wang, Solar-Lezama, Sen, and Stoica]{jain2024livecodebench}
Naman Jain, King Han, Alex Gu, Wen-Ding Li, Fanjia Yan, Tianjun Zhang, Sida Wang, Armando Solar-Lezama, Koushik Sen, and Ion Stoica.
\newblock Livecodebench: Holistic and contamination free evaluation of large language models for code.
\newblock \emph{arXiv preprint arXiv:2403.07974}, 2024.

\bibitem[Li et~al.(2023)Li, Zhang, Koto, Yang, Zhao, Gong, Duan, and Baldwin]{li2023cmmlu}
Haonan Li, Yixuan Zhang, Fajri Koto, Yifei Yang, Hai Zhao, Yeyun Gong, Nan Duan, and Timothy Baldwin.
\newblock Cmmlu: Measuring massive multitask language understanding in chinese.
\newblock \emph{arXiv preprint arXiv:2306.09212}, 2023.

\bibitem[Li et~al.(2025)Li, Su, Yang, Xie, Wang, Xie, Wong, and Yang]{li2025quantization}
Zhen Li, Yupeng Su, Runming Yang, Congkai Xie, Zheng Wang, Zhongwei Xie, Ngai Wong, and Hongxia Yang.
\newblock Quantization meets reasoning: Exploring llm low-bit quantization degradation for mathematical reasoning.
\newblock \emph{arXiv preprint arXiv:2501.03035}, 2025.

\bibitem[Lightman et~al.(2023)Lightman, Kosaraju, Burda, Edwards, Baker, Lee, Leike, Schulman, Sutskever, and Cobbe]{lightman2023let}
Hunter Lightman, Vineet Kosaraju, Yuri Burda, Harrison Edwards, Bowen Baker, Teddy Lee, Jan Leike, John Schulman, Ilya Sutskever, and Karl Cobbe.
\newblock Let's verify step by step.
\newblock In \emph{The Twelfth International Conference on Learning Representations}, 2023.

\bibitem[Liu et~al.(2024)Liu, Feng, Xue, Wang, Wu, Lu, Zhao, Deng, Zhang, Ruan, et~al.]{liu2024deepseek}
Aixin Liu, Bei Feng, Bing Xue, Bingxuan Wang, Bochao Wu, Chengda Lu, Chenggang Zhao, Chengqi Deng, Chenyu Zhang, Chong Ruan, et~al.
\newblock Deepseek-v3 technical report.
\newblock \emph{arXiv preprint arXiv:2412.19437}, 2024.

\bibitem[Liu et~al.(2023{\natexlab{a}})Liu, Xia, Wang, and Zhang]{liu2023your}
Jiawei Liu, Chunqiu~Steven Xia, Yuyao Wang, and Lingming Zhang.
\newblock Is your code generated by chatgpt really correct? rigorous evaluation of large language models for code generation.
\newblock \emph{Advances in Neural Information Processing Systems}, 36:\penalty0 21558--21572, 2023{\natexlab{a}}.

\bibitem[Liu et~al.(2025)Liu, Sun, Zhang, Bai, Yu, Yu, Yuan, and Hou]{liu2025quantization}
Ruikang Liu, Yuxuan Sun, Manyi Zhang, Haoli Bai, Xianzhi Yu, Tiezheng Yu, Chun Yuan, and Lu~Hou.
\newblock Quantization hurts reasoning? an empirical study on quantized reasoning models.
\newblock \emph{arXiv preprint arXiv:2504.04823}, 2025.

\bibitem[Liu et~al.(2023{\natexlab{b}})Liu, Oguz, Zhao, Chang, Stock, Mehdad, Shi, Krishnamoorthi, and Chandra]{liu2023llm}
Zechun Liu, Barlas Oguz, Changsheng Zhao, Ernie Chang, Pierre Stock, Yashar Mehdad, Yangyang Shi, Raghuraman Krishnamoorthi, and Vikas Chandra.
\newblock Llm-qat: Data-free quantization aware training for large language models.
\newblock \emph{arXiv preprint arXiv:2305.17888}, 2023{\natexlab{b}}.

\bibitem[MAA(2024)]{AIME2024}
MAA.
\newblock American invitational mathematics examination - aime.
\newblock In \emph{American Invitational Mathematics Examination - AIME 2024}, February 2024.
\newblock URL \url{https://maa.org/math-competitions/american-invitational-mathematics-examination-aime}.

\bibitem[Min et~al.(2024)Min, Chen, Jiang, Chen, Deng, Hu, Tang, Wang, Cheng, Song, et~al.]{min2024imitate}
Yingqian Min, Zhipeng Chen, Jinhao Jiang, Jie Chen, Jia Deng, Yiwen Hu, Yiru Tang, Jiapeng Wang, Xiaoxue Cheng, Huatong Song, et~al.
\newblock Imitate, explore, and self-improve: A reproduction report on slow-thinking reasoning systems.
\newblock \emph{arXiv preprint arXiv:2412.09413}, 2024.

\bibitem[Nyamsuren(2024)]{nyamsuren2024evaluating}
Enkhbold Nyamsuren.
\newblock Evaluating quantized large language models for code generation on low-resource language benchmarks.
\newblock \emph{arXiv preprint arXiv:2410.14766}, 2024.

\bibitem[Qin et~al.(2024)Qin, Li, Zou, Liu, Xia, Huang, Ye, Yuan, Liu, Li, et~al.]{qin2024o1}
Yiwei Qin, Xuefeng Li, Haoyang Zou, Yixiu Liu, Shijie Xia, Zhen Huang, Yixin Ye, Weizhe Yuan, Hector Liu, Yuanzhi Li, et~al.
\newblock O1 replication journey: A strategic progress report--part 1.
\newblock \emph{arXiv preprint arXiv:2410.18982}, 2024.

\bibitem[Rein et~al.(2024)Rein, Hou, Stickland, Petty, Pang, Dirani, Michael, and Bowman]{rein2024gpqa}
David Rein, Betty~Li Hou, Asa~Cooper Stickland, Jackson Petty, Richard~Yuanzhe Pang, Julien Dirani, Julian Michael, and Samuel~R Bowman.
\newblock Gpqa: A graduate-level google-proof q\&a benchmark.
\newblock In \emph{First Conference on Language Modeling}, 2024.

\bibitem[Romero et~al.(2014)Romero, Ballas, Kahou, Chassang, Gatta, and Bengio]{romero2014fitnets}
Adriana Romero, Nicolas Ballas, Samira~Ebrahimi Kahou, Antoine Chassang, Carlo Gatta, and Yoshua Bengio.
\newblock Fitnets: Hints for thin deep nets.
\newblock \emph{arXiv preprint arXiv:1412.6550}, 2014.

\bibitem[Shen et~al.(2024)Shen, Kong, Yang, Han, Lu, Dong, Lyu, Li, Guo, Shu, et~al.]{shen2024edgeqat}
Xuan Shen, Zhenglun Kong, Changdi Yang, Zhaoyang Han, Lei Lu, Peiyan Dong, Cheng Lyu, Chih-hsiang Li, Xuehang Guo, Zhihao Shu, et~al.
\newblock Edgeqat: Entropy and distribution guided quantization-aware training for the acceleration of lightweight llms on the edge.
\newblock \emph{arXiv preprint arXiv:2402.10787}, 2024.

\bibitem[Xiao et~al.(2023)Xiao, Lin, Seznec, Wu, Demouth, and Han]{xiao2023smoothquant}
Guangxuan Xiao, Ji~Lin, Mickael Seznec, Hao Wu, Julien Demouth, and Song Han.
\newblock Smoothquant: Accurate and efficient post-training quantization for large language models.
\newblock In \emph{International Conference on Machine Learning}, pp.\  38087--38099. PMLR, 2023.

\bibitem[Xu et~al.(2024)Xu, Li, Tao, Shen, Cheng, Li, Xu, Tao, and Zhou]{xu2024survey}
Xiaohan Xu, Ming Li, Chongyang Tao, Tao Shen, Reynold Cheng, Jinyang Li, Can Xu, Dacheng Tao, and Tianyi Zhou.
\newblock A survey on knowledge distillation of large language models.
\newblock \emph{arXiv preprint arXiv:2402.13116}, 2024.

\bibitem[Yang et~al.(2024)Yang, Zhu, Lu, Wang, Chen, Gao, Yan, and Chen]{yang2024survey}
Chuanpeng Yang, Yao Zhu, Wang Lu, Yidong Wang, Qian Chen, Chenlong Gao, Bingjie Yan, and Yiqiang Chen.
\newblock Survey on knowledge distillation for large language models: methods, evaluation, and application.
\newblock \emph{ACM Transactions on Intelligent Systems and Technology}, 2024.

\bibitem[Yu et~al.(2024)Yu, Wang, Shan, and Wan]{yu2024super}
Mengxia Yu, De~Wang, Qi~Shan, and Alvin Wan.
\newblock The super weight in large language models.
\newblock \emph{arXiv preprint arXiv:2411.07191}, 2024.

\end{thebibliography}
\bibliographystyle{colm2024_conference}

\appendix
\newpage
\section{Additional details}

\subsection{Quantization Implementation Details}
\label{appendix:1}

\begin{table*}[!ht]
\centering
\renewcommand{\arraystretch}{1.8} 
\begin{tabular}{@{}lcccccc@{}}
\toprule

\makecell{\textbf{Weight-Matrix} } & 
\makecell{\textbf{Q4\_K\_M}} & 
\makecell{\textbf{Q3\_K\_M}} & 
\makecell{\textbf{DQ3\_K\_M (ours)}} & 
\makecell{\textbf{Q2\_K\_L}} & 
\makecell{\textbf{Q2\_K\_XL}} & 

\\

\midrule

\textbf{output} & q6\_k & q6\_k & q6\_k & q6\_k & q6\_k   \\
\textbf{token\_embd} & q4\_k & q3\_k & q4\_k & q4\_k & q4\_k   \\
\textbf{attn\_kv\_a\_mqa} & q4\_k & q3\_k & q6\_k & q6\_k & q6\_k   \\
\textbf{attn\_kv\_b} & q4\_k & q3\_k & q6\_k & q2\_k & q6\_k   \\
\textbf{attn\_output} & q4\_k & q4\_k & q4\_k & q3\_k & q4\_k   \\
\textbf{attn\_q\_a} & q4\_k & q3\_k & q4\_k & q2\_k & q4\_k   \\
\textbf{attn\_q\_b} & q4\_k & q3\_k & q4\_k & q2\_k & q4\_k   \\
\textbf{ffn\_down} & q6\_k & q5\_k & q6\_k & q3\_k & q6\_k   \\
\textbf{ffn\_gate} & q4\_k & q3\_k & q4\_k & q2\_k & q4\_k   \\
\textbf{ffn\_up} & q4\_k & q3\_k & q4\_k & q2\_k & q4\_k   \\

\addlinespace[0.4em]

\textbf{ffn\_down\_exps} & \makecell[c]{ q4\_k(53.4\%) \\  q6\_k(46.6\%)}  & q4\_k &  \makecell[c]{ q3\_k(75.9\%) \\  q4\_k(20.7\%) \\  q6\_k(3.40\%)}   &  q3\_k &  \makecell[c]{ q2\_k(94.8\%) \\  q3\_k(5.20\%)} \\

\addlinespace[0.4em]

\textbf{ffn\_down\_shexp} &  \makecell[c]{ q4\_k(53.4\%) \\  q6\_k(46.6\%)} & q4\_k & q6\_k & q3\_k & q6\_k   \\

\addlinespace[0.4em]

\textbf{ffn\_gate\_exp} & q4\_k & q3\_k & q3\_k & q2\_k & q2\_k   \\
\textbf{ffn\_gate\_shexp} & q4\_k & q3\_k & q4\_k & q2\_k & q4\_k   \\
\textbf{ffn\_up\_exps} & q4\_k & q3\_k & q3\_k & q2\_k & q2\_k   \\
\textbf{ffn\_up\_shexp} & q4\_k & q3\_k & q4\_k & q2\_k & q4\_k   \\

\bottomrule
\end{tabular}
\vspace{10pt} %
\caption{Quantization implementation details of different methods}
\label{tab:quantization_details}
\end{table*}

\newpage

\subsection{Benchmark Statistics}
\label{appendix:2}

\begin{table*}[!ht]
\centering
\renewcommand{\arraystretch}{1.8} 
\begin{tabular}{@{}lcccccc@{}}
\toprule

\makecell{\textbf{Benchmark} } & 
\makecell{\textbf{Question Count}} &
\makecell{\textbf{Weight} } \\

\midrule

\textbf{AIME 2024} & 30 & 0.2 \\
\textbf{MATH 500} & 500 & 0.5  \\
\textbf{GPQA} & 198 & 0.5 \\
\textbf{MBPP} &  378 & 0.5 \\
\textbf{MBPP+} &  378 & 0.5 \\
\textbf{LiveCodeBench} & 272 & 0.5 \\
\textbf{MMLU} & 14042 & 1 \\
\textbf{CMMLU} & 11582 & 1 \\
\textbf{C-Eval} & 12342 & 1 \\

\bottomrule
\end{tabular}
\vspace{10pt} %
\caption{The statistics of benchmarks for evaluation (The weight is used for calculating weighted average scores in experiments.)}
\label{tab:benchmark}
\end{table*}

\end{document}